\documentclass[conference]{IEEEtran}
\IEEEoverridecommandlockouts
\usepackage{cite}
\usepackage{amsmath,amssymb,amsfonts}
\usepackage{algorithmic}
\usepackage{graphicx}
\usepackage{textcomp}
\usepackage{xcolor}
\def\BibTeX{{\rm B\kern-.05em{\sc i\kern-.025em b}\kern-.08em
    T\kern-.1667em\lower.7ex\hbox{E}\kern-.125emX}}
\begin{document}

\title{Time CNN and Graph Convolution Network for Epileptic Spike Detection in MEG Data
}

\author{\IEEEauthorblockN{Pauline Mouches}
\IEEEauthorblockA{\textit{Lyon Neuroscience Research center, INSERM U1028} \\
Lyon, France\\
pauline.mouches@inserm.fr}
\and
\IEEEauthorblockN{Thibaut Dejean}
\IEEEauthorblockA{\textit{Lyon Neuroscience Research center, INSERM U1028} \\
Lyon, France}
\and
\IEEEauthorblockN{Julien Jung}
\IEEEauthorblockA{\textit{Lyon Neuroscience Research center, INSERM U1028} \\
\textit{Functional Neurology and Epileptogy, HCL}\\
Lyon, France}
\and
\IEEEauthorblockN{Romain Bouet}
\IEEEauthorblockA{\textit{Lyon Neuroscience Research center, INSERM U1028} \\
Lyon, France}
\and
\IEEEauthorblockN{Carole Lartizien}
\IEEEauthorblockA{\textit{Univ Lyon, CNRS, Inserm, INSA Lyon, UCBL, CREATIS} \\
\textit{UMR5220, U1206, F-69621}\\
Villeurbanne, France}
\and
\IEEEauthorblockN{Romain Quentin}
\IEEEauthorblockA{\textit{Lyon Neuroscience Research center, INSERM U1028} \\
Lyon, France}
}

\maketitle

\begin{abstract}
Magnetoencephalography (MEG) recordings of patients with epilepsy exhibit spikes, a typical biomarker of the pathology. Detecting those spikes allows accurate localization of brain regions triggering seizures. Spike detection is often performed manually. However, it is a burdensome and error prone task due to the complexity of MEG data. To address this problem, we propose a 1D temporal convolutional neural network (Time CNN) coupled with a graph convolutional network (GCN) to classify short time frames of MEG recording as containing a spike or not. Compared to other recent approaches, our models have fewer parameters to train and we propose to use a GCN to account for MEG sensors spatial relationships. Our models produce clinically relevant results and outperform deep learning-based state-of-the-art methods reaching a classification f1-score of 76.7\% on a balanced dataset and of 25.5\% on a realistic, highly imbalanced dataset, for the spike class.
\end{abstract}


\section{Introduction}
Magnetoencephalography (MEG) is a non-invasive technique to record neural activity with a high temporal and spatial resolution. A major clinical application of MEG is for the pre-surgical evaluation of drug-resistant epilepsy to identify brain regions triggering seizures \cite{b1}. MEG recordings of epilepsy patients present, outside seizures, morphologically defined events, called interictal spikes. These spikes are focal, visible only on a subset of MEG sensors, and highly variable among patients \cite{b1}. Their identification in the MEG recordings allows accurate localization of brain regions triggering seizures, but is a burdensome task due to the large number of sensors.

\begin{figure*}[htbp]
\centerline{\includegraphics[scale=0.35]{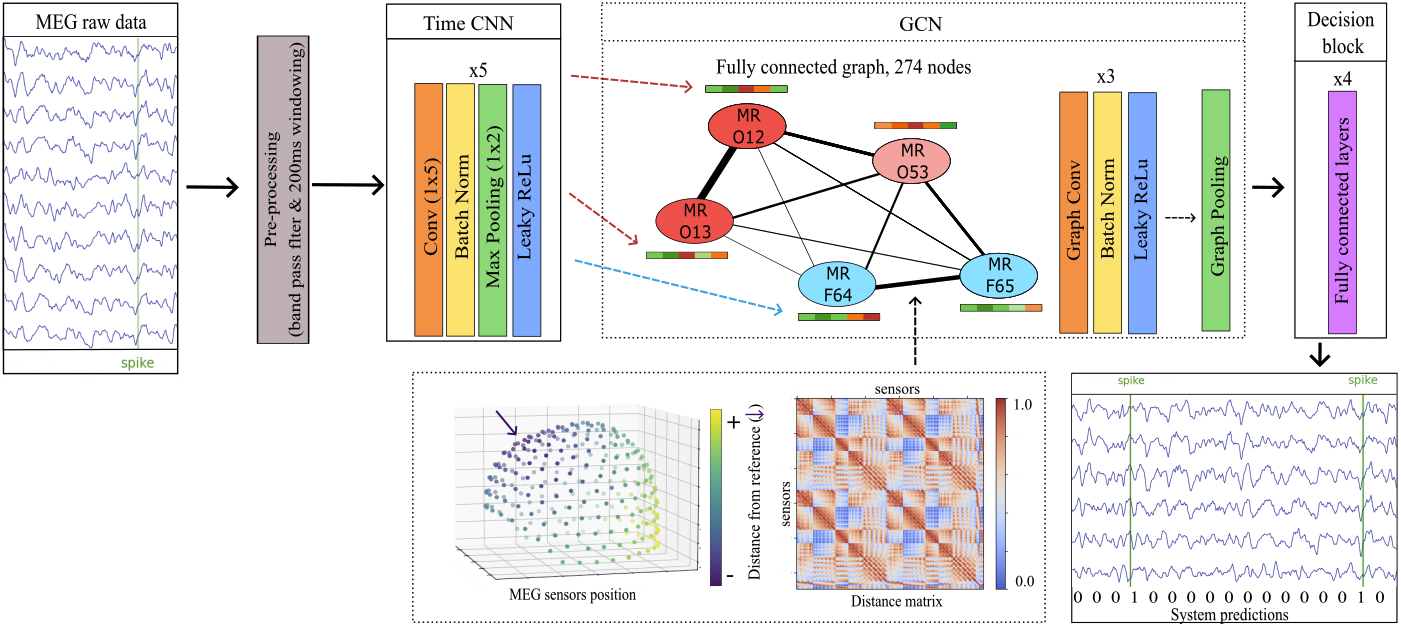}}
\caption{Graphical diagram of the proposed approach. Raw MEG data with annotated spike timings are used as input. Plain line boxes illustrate the Time convolutional neural network (CNN) model architecture while dashed line boxes and arrows show the graph convolutional network (GCN) components. In the GCN, sensor-specific features from the Time CNN are used as node features and the normalized geodesic distance between sensors is used to compute edge weights. The decision block outputs a binary prediction for each short MEG time frame.}
\label{fig1}
\end{figure*}

For these reasons, automated methods for interictal spike detection were proposed. Most methods were based on traditional machine learning models and consisted in extracting features from short time frames of MEG recordings and then using a classifier to identify frames with spikes \cite{b2}. Recently, two studies developed deep learning methods, offering more generalizable and better optimized models compared to those based on a priori features \cite{b3,b4}. Both proposed to use deep convolutional neural network (CNNs) applied on time frames of raw MEG data, as shown on the left box of Fig. \ref{fig1}.

Zeng et al. \cite{b3} proposed a model called EMS-Net, in which sensors are split into 39 subgroups based on their spatial proximity, corresponding to different brain cortical regions. The authors annotated short MEG time frames (300ms) per subgroup and implemented a CNN extracting both single-sensor features and global features, from the subgroup of sensors. The single-sensor branch of the CNN uses 1D convolutions over time while the global branch uses 2D convolutions over the MEG time frame. Both feature types are combined for the final decision. The authors tested their method on recordings from 22 patients. They demonstrated increased classification performances compared to traditional machine learning methods including a support vector machine model and a logistic regression model trained on the flattened data from the time frames. Nevertheless, they tackled a simplified problem by looking at sensors’ subgroups. Training such a model requires intensive data annotating to get a label per sensors subgroup, which is very rare in clinical practice where simpler annotations containing spike timing information only are most often realized. 

Hirano et al. \cite{b4} used a large database of 375 annotated MEG recordings to train and evaluate a 2D CNN, called FAMED, based on a SE-ResNet \cite{b5} to classify MEG time frames of 2 seconds. Then, the authors developed a segmentation model to delineate the spike within the frame. When comparing their classification model to EMS-Net, the authors found higher classification accuracy with their approach, although they specify that only typical spikes were selected for the task, which might not be representative of realistic data.

None of the previously proposed approaches accounted for 3D spatial relationships between MEG sensors within the classification model. However, this information was shown to be relevant in other medical applications based on similar recordings where graph-based models were used to integrate sensor relationship information \cite{b6}. In \cite{b7}, for example, a graph convolutional network (GCN) was used to distinguish normal from pathological electroencephalography (EEG) recording time frames of 30 seconds using frequency features extracted as a prior step. In this context, the proposed GCN outperformed two comparison models: a fully-connected neural network and a random forest model trained on the same frequency features. Nevertheless, to our knowledge, no previous study considered the use of GCNs in the context of interictal spike detection \cite{b6}.

In this work, we present a lightweight Time CNN model coupled with a GCN for automated interictal spike detection in raw MEG recordings (Fig. 1). Our method aims to classify short MEG recording frames as containing an interictal spike or not. The lightweight Time CNN model extracts sensor-specific temporal features with 1D convolutional kernels. Coupling our Time CNN with a GCN (Time CNN-GCN model) allows to model the distances between MEG sensors. Our main contributions are (1) a lightweight Time CNN model which outperforms deeper CNN models for automated spike detection in a balanced and realistic dataset and (2) the investigation of the benefits of using a GCN to account for MEG sensors spatial relationships. We test and compare our models to two state-of-the-art deep learning models \cite{b3,b4}  on MEG recordings from a large cohort of 95 epileptic patients. 

\section{Proposed Method}

\subsection{Problem definition}

In this work, the aim is to train a binary classification model on short time frames of MEG recordings, labeled as containing a spike or not. 

Each time frame can be denoted as $X_{i} \in R^{(ns \times nt)}$ where $ns$ corresponds to the number of MEG sensors $S=\left\{S_{1},...,S_{ns}\right\}$ and $nt$ to the number of time points. 

\subsection{Time CNN}

The Time CNN model extracts sensor-specific temporal features using 1D convolution kernels. Its architecture was inspired from the work by Peng et al. \cite{b8}. It is made of five blocks,  each one containing a convolutional layer with a (1 $\times$ 5) kernel and varying number of channels (32, 64, 128, 256, 1) followed by a batch normalization layer and leaky ReLU activation. The first two blocks include a max pooling layer with a 1D kernel (1x2) to reduce the feature space dimensionality. The 1D kernels ensure that no information leaks between the sensors. The output of the CNN is a feature matrix $Z_{i} \in R^{(ns \times nf)}$ where $nf$ represents the number of temporal features for each sensor, i.e., 7. Finally, the feature matrix is flattened and passed through the decision block made of four fully connected layers of size (128, 64, 32, 1), the last one having a sigmoid activation.

\subsection{Coupling with a GCN}

In a second time, we propose to insert a GCN between the convolutional layers and the fully connected layers of the decision block. To do so, after the final convolutional layer, the samples $Z_{i}$ are represented as graphs $G_{i}=(V,E)$ where nodes are the set of MEG sensors with $|V|=ns$ and $E$ the set of edges with $|E|=\frac{ns \times (ns-1)}{2}$ as we have an undirected fully connected graph. $E$ is represented as an adjacency matrix $A \in R^{(ns \times ns)}$ with $A_{j,k}=1-d(S_{j}, S_{k})$, where $d$ is the min-max normalized geodesic distance. Therefore, edges connecting neighbouring nodes have high weights. Finally, the output of the CNN, $Z_{i}$, represents the node feature vectors of $G_{i}$. GCN layers \cite{b9} perform convolutions over the graph nodes resulting in updated node features depending on neighboring node features, edge weights and trainable parameters. The GCN contains three graph convolutional layers with hidden dimensions of (30, 128, 256), followed by a global add pooling layer, which adds graph node features across nodes, resulting in one embedding vector of length 256 representing the full graph $G_{i}$. This graph embedding is finally passed through the decision block as described in section II. B.

\section{Experiments and Results}

\subsection{Data and preprocessing}

The dataset is composed of MEG recordings of 95 epileptic patients of around 9 minutes each. Recordings were acquired on a CTF Omega system with 274 sensors. The records of each patient have been analyzed and annotated by an expert neurologist who identified interictal spike timings. 

As preprocessing steps, recordings were bandpass filtered (0.5-50Hz) and resampled to 150Hz to ensure that any low-frequency drift or interference is removed from the signal and allowing us to focus on the frequency range that is most pertinent to detect interictal spike. Frames of MEG recordings of 200ms in length, corresponding to 30 time points, were created. By using 200ms frames, we can accurately capture the characteristics of individual spikes, which duration is approximately 80ms, without losing important information. Moreover, frames are defined such that they overlap of 60ms and frames with spikes located on the border of the frame ($<$30ms from the border) are considered as spike-free frames. 
Over the 95 patients, the number of spike positive frames was 4584 and that of spike negative frames was 360785, which corresponds to a frame displaying a spike out of $\sim$80 frames. Data preprocessing was performed using mne-python \cite{b10}.

\subsection{Model training}

For the model training, we created an artificially balanced dataset by randomly selecting as many negative frames as available spike positive frames (around 4600). This allowed us to significantly reduce the model’s training time.

Five repetitions of 10-folds cross-validation iterations were employed. For each repetition, folds were split across patients which ensures that frames from a specific patient are not spread across folds. Within each cross-validation iteration, 10\% of the training patients were assigned to the validation set.

The training parameters for our model are as follows: weight initialization using the Xavier method \cite{b11}, binary cross-entropy loss function, a maximum number of training epochs   set at 50 with early stopping by monitoring the validation loss. We used the Adam optimizer \cite{b12} with a learning rate of 1e-3 and a batch size of 32. To further prevent overfitting, dropout layers with a probability of 0.3 were used between each fully connected layer of the decision block. The model was implemented using Pytorch \cite{b13} and Pytorch Geometric \cite{b14}. Finally, the implementations of the comparison models were taken from the code made available by Hirano et al. \cite{b4}. For the training of EMS-Net, no sensor subgroups were used but rather the full list of sensors as no sensor-specific annotations are available in our data.

\subsection{Results on balanced test data}

To test the models on balanced data, we selected as many negative frames as available spike positive frames within the test dataset.

\begin{table}[htbp]
\caption{Classification performances on a balanced test dataset (\%). Metrics correspond to the average and standard deviation across the 5 repetitions of 10 folds cross-validation. F1-score, Specificity, and Sensitivity are for the spike class.}
\begin{center}
\begin{tabular}{|p{1.5cm}|c|c|c|c|}
\hline
\textbf{Model}&\textbf{Accuracy}&\textbf{F1-score}&\textbf{Specificity}&\textbf{Sensitivity} \\
\hline
\textbf{FAMED}&66.8$\pm$1.3&	66.3$\pm$0.7&67.2$\pm$5.5&66.4$\pm$3.1\\
\hline
\textbf{EMS-Net}&72.9$\pm$1.2&69.8$\pm$2.0&81.0$\pm$1.4&64.8$\pm$3.3\\
\hline
\textbf{Time CNN} \textbf{(Ours)}&
76.9$\pm$0.7&75.4$\pm$1.0&\textbf{82.0$\pm$1.2}&71.9$\pm$1.7
\\
\hline
\textbf{Time CNN-GCN}\textbf{(Ours)}&\textbf{77.5$\pm$0.6}&\textbf{76.7$\pm$0.5}&80.0$\pm$1.6&\textbf{75.0$\pm$0.8}\\
\hline
\end{tabular}
\label{tab1}
\end{center}
\end{table}

Table \ref{tab1} shows result metrics of our models (Time CNN and Time CNN-GCN) and state-of-the-art models on a balanced test dataset. The Time CNN-GCN model outperforms other models with a classification accuracy of 77.5\% and a f1-score of 76.7\% for the spike class, slightly higher than the Time CNN only (f1-score: 75.4\%). The major difference between our two models is observed on the sensitivity score, which is higher for the Time CNN-GCN (3.1\% difference), indicating its ability to detect more spikes than the Time CNN only. Our Time CNN-GCN sensitivity is also higher than the sensitivity of state-of the art models. This result might be explained by the fact that the graph edges bring complementary information allowing to detect more spikes that might be less visible or of lower amplitude.  

For state-of-the-art models, EMS-Net also demonstrates good performances (accuracy: 72.9\%) with a specificity of 81\% but lower sensitivity (64.8\%), indicating more undetected spikes. As EMS-Net is deeper than our proposed model, its training might be less optimal than simpler models. The FAMED model leads to worse performances (accuracy: 66.8\%, f1-score: 66.3\%). This could be explained by the fact that the FAMED model architecture was initially designed to classify large frames ($\sim$2 seconds), and might not be optimal for our problem with smaller frames (200ms). Finally, we observe lower standard deviation across the 5 repetitions of 10-fold cross-validation for our models, compared to the state-of-the-art models, which shows higher stability for our models.

\subsection{Results on imbalanced, realistic test data}

Our ultimate goal is to help clinicians by automatically detecting interictal spikes on minimally preprocessed data and hence imbalanced data. We evaluated the models trained on a balanced dataset on imbalanced data. To do so, for each cross-validation iteration, all frames from the patients from the test fold are now used to test the model. A threshold moving strategy is then employed, as done by Zheng et al. \cite{b3}, to overcome the differences in data distribution between the training data (balanced) and testing data (imbalanced). This method consists in increasing the threshold to binarize the model probability outputs, which is typically set to 0.5 in a balanced data setting. Thus, the optimal threshold for each model was found by maximizing the f1-score for the spike class on all frames of patients from the validation set. This optimal threshold was then applied to binarize the output probabilities of the test set.

\begin{table}[htbp]
\caption{Classification performances on an imbalanced test dataset (\%). Metrics correspond to the average and standard deviation across the 5 repetitions of 10 folds cross-validation. F1-score, Specificity, and Sensitivity are for the spike class.}
\begin{center}
\begin{tabular}{|p{1.5cm}|c|c|c|c|}
\hline
\textbf{Model}&\textbf{Accuracy}&\textbf{F1-score}&\textbf{Specificity}&\textbf{Sensitivity} \\
\hline
\textbf{FAMED}&96.9$\pm$0.4&12.0$\pm$1.4&97.9$\pm$0.4&16.7 $\pm$1.1\\
\hline
\textbf{EMS-Net}&96.6$\pm$0.6&17.1$\pm$2.2&97.5$\pm$0.7&27.2$\pm$2.9\\
\hline
\textbf{Time CNN}\textbf{(Ours)} & \textbf{97.9$\pm$0.3} & \textbf{25.5$\pm$2.3} & \textbf{98.8$\pm$0.3} & 28.5$\pm$2.4\\
\hline
\textbf{Time CNN-GCN}\textbf{(Ours)}&  96.9$\pm$0.5 & 20.2$\pm$1.4 & 97.7$\pm$0.5 & \textbf{31.5$\pm$3.7}\\
\hline
\end{tabular}
\label{tab2}
\end{center}
\end{table}

\begin{figure}[htbp]
\centerline{\includegraphics[scale=0.17]{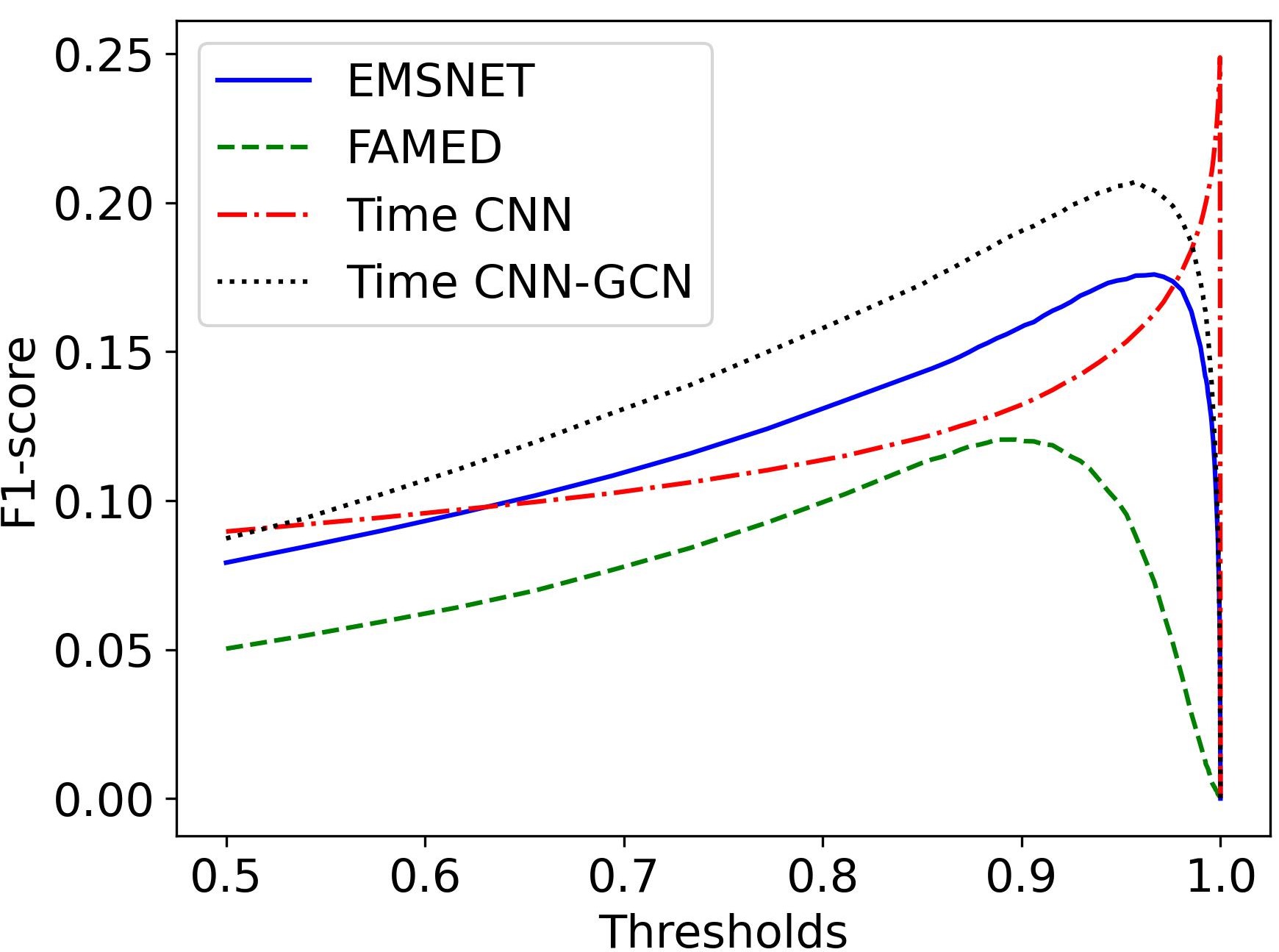}}
\caption{Averaged F1-scores across the 5 repeated 10 folds cross-validation iterations for different thresholds to binarize model outputs on the imbalanced dataset.}
\label{fig2}
\end{figure}

Table \ref{tab2} shows performance metrics obtained with the optimal threshold for each model. Furthermore, Fig. \ref{fig2} illustrates the f1-score of each model as a function of the threshold. For all models, optimal thresholds are above 0.9. We observe an expected drop of performance compared with the results on balanced test data (Table \ref{tab1}). Among the models, both of our models outperform state-the-art-models, with the Time CNN reaching the highest f1-score (25.5\%), while EMS-Net and FAMED reach maximum f1-scores of 17.1\% and 12.0\%, respectively.

Second, we observe that, when optimizing the threshold on the validation data, the Time CNN alone performs better than when coupled with the GCN, although the opposite was observed on a balanced test dataset. Indeed, the Time CNN-GCN tends to detect more spikes also leading to more false positives, reflected by the lower specificity. Interestingly, Fig. \ref{fig2} shows that the f1-score of the Time CNN remains low, around 0.1, and sharply increases for very high thresholds. This indicates that the Time CNN tends to predict extreme probabilities (red curve) compared to the Time CNN-GCN, which produces smoother probability outputs (black curve). Obtaining such smooth probabilities is potentially more clinically relevant as it could allow better artefact identification. Finally, the threshold moving technique might not be efficient and more advanced methods should be considered to handle the imbalanced nature of the data.

\section{Discussion and Conclusion}

In this paper, we proposed novel approaches to analyze raw MEG data of epileptic patients. First, our Time CNN demonstrates better results than more complex/deeper models proposed in the past. Although this finding might be counterintuitive, the superiority of simple models in the context of medical data analysis has already been demonstrated on different data type such as chest X-ray images \cite{b15} or brain MRI \cite{b8}. Future work investigating the model internal features should be carried out to understand better this observation. 
	
Then, coupling the Time CNN with a GCN leads to modest model performance improvements on balanced test data. The GCN brings complementary information about sensors spatial localization, which leads to an increased sensitivity. As more spikes are detected, more false positives are also observed. Therefore, providing sensors distances as graph edge weights might not be optimal and other information such as spectral coherence between sensor data, as done in \cite{b7}, should be investigated in future work. Moreover, including more patient data could improve the Time CNN-GCN model training as it would bring more diversity in spike spatial localization.
	
Finally, although our models outperform state-of-the-art approaches on the imbalanced test data, our results remain moderate. Using a balanced dataset simplifies model training but the threshold moving approach does not allow us to compensate for the high imbalance ratio. Perspective of improvements include finding a better adapted approach for model training on the realistic imbalanced dataset with an adjusted loss function, such as a focal loss \cite{b16}.

From a clinical perspective, our models demonstrate promising results for interictal spike detection from electrophysiological recordings. In terms of technical improvements, we proposed an innovative approach which could be transferred to other applications for MEG data analysis.

\section*{Acknowledgment}

This work was supported by a postdoctoral fellowship from the Fondation pour la Recherche Médicale and by the ATIP Avenir program.

\section*{Compliance with Ethical Standards}

Ethic Committee agreement was obtained for the study (CCP Lyon Sud Est IV Research Ethics Committee, approval number: N2012-A00516-37, 05/24/2012) in accordance with the Declaration of Helsinki. Patients were fully informed and signed a written consent form

\end{document}